\documentclass[letterpaper, 10 pt, conference]{ieeeconf}  

\IEEEoverridecommandlockouts                              

\usepackage{graphics}
\usepackage{epsfig}
\usepackage{mathptmx}
\usepackage{times}
\usepackage{amsmath} 
\usepackage{amssymb}  
\usepackage{xcolor}
\usepackage{url}
\usepackage{booktabs}
\usepackage[T1]{fontenc}
\usepackage{threeparttable}
\usepackage{multirow}
\usepackage{pifont}
\usepackage{listings}
\usepackage{tikz}
\usepackage{float}

\usetikzlibrary{positioning,arrows.meta,fit,backgrounds,calc,shapes.geometric}

\newcommand{\cmark}{\ding{51}}
\newcommand{\xmark}{\ding{55}}

\newcommand{\up}{{\tiny$\uparrow$}}
\newcommand{\down}{{\tiny$\downarrow$}}

\title{\LARGE \bf
ShelfChange3D: Object-Level 3D Change Detection for Retail Shelf Monitoring}

\author{
Lingyi Zhou$^{1}$,
Yunke Wang$^{1}$,
Mengyu Zheng$^{2}$,
Wenbo Wang$^{1}$,
Zijian Wang$^{3}$,
Chang Xu$^{1}$\\
\small $^{1}$The University of Sydney \qquad
$^{2}$Beijing Jiaotong University \qquad
$^{3}$StellarEdge AI
}

\begin{document}

\maketitle
\thispagestyle{empty}
\pagestyle{empty}

\begin{abstract}

Reliable shelf monitoring is an important capability for retail automation, yet existing out-of-stock detection methods mainly operate in image space and lack metric 3D localization for downstream robotic systems. We formulate shelf monitoring as object-level 3D change detection: given two RGB-D observations captured at different times, the goal is to identify changed products and localize each change with a 3D bounding box. To support this task, we introduce ShelfChange3D, comprising 145K synthetic and 5K real-world paired RGB-D observations with object-level 3D change annotations. We further propose ChangeBox, an end-to-end framework that jointly reasons over paired observations and predicts object-level 3D change boxes. To improve localization accuracy, we introduce a geometry-based refinement stage that exploits depth and gravity prior to estimate relative pose and refine predicted boxes. Experiments show that ChangeBox outperforms existing change detection baselines, with further gains from refinement and effective transfer from synthetic to real-world observations. 
Our code will be available at our project website: \url{https://zerone0011.github.io/ShelfChange3D/}.

\end{abstract}
\section{INTRODUCTION}
\label{sec:intro}

Automated shelf monitoring is an important component of retail automation, enabling stores and robotic systems to track shelf conditions and respond to inventory changes in a timely manner. Retail shelves change constantly as customers take products and staff restock them, and products that sell out without timely replenishment can lead to lost sales and poor customer experience. Out-of-stock (OOS) detection therefore aims to automatically identify missing products from shelf observations, reducing the need for frequent manual inspection. For robotic retail systems, however, detecting that a product is missing is only part of the problem: the system must also determine where the change occurred in the physical environment to support subsequent actions such as inspection or replenishment.

Existing vision-based OOS methods mainly operate in image space, using attention maps, segmentation, or 2D object detection~\cite{allegra2021exploiting,santra2022graph,de2023autonomous,vsikic2024enhanced}. Existing retail datasets are similarly designed around 2D annotations, such as bounding boxes, points, or vacancy labels~\cite{goldman2019precise,allegra2021exploiting,out-of-stock-b2irt_dataset,empty-slots-in-shelves_dataset}. Although depth information has been explored for estimating shelf occupancy~\cite{milella2020towards}, existing approaches do not explicitly localize product-level changes in metric 3D space. Consequently, their outputs remain tied to the camera view and provide neither the 3D position nor the spatial extent of the changed region, limiting their direct use for downstream robotic perception and action.

This limitation motivates the formulation of shelf monitoring as an object-level 3D change detection problem: given two shelf observations captured at different times, the goal is to identify what has changed and localize each changed product in 3D. 
Existing 3D change detection methods identify changed points, regions, or objects by comparing 3D observations over time~\cite{adam2022objects,adam2023has,zhou20263d}, but these methods generally do not localize individual changes with 3D bounding boxes. Standard 3D object detection, on the other hand, provides object-level 3D localization~\cite{qi2019deep,Misra2021AnET,wang2022detr3d}, but operates on a single observation and detects objects that are currently present. In OOS monitoring, however, the target product may have already disappeared from the current observation. Shelf monitoring therefore requires both cross-time change reasoning and object-level 3D localization.

To address this problem, we introduce \textbf{ShelfChange3D}, a benchmark of 145K synthetic and 5K real-world paired RGB-D shelf observations with object-level 3D change annotations, and develop \textbf{ChangeBox}, an end-to-end framework that jointly reasons over two observations and predicts changed products as 3D bounding boxes. Our analysis shows that, while ChangeBox can reliably identify product-level changes, accurate 3D localization remains challenging, with box center estimation emerging as the dominant source of error. Motivated by this observation, we further introduce a geometry-based refinement stage that exploits depth and a gravity prior to estimate the relative pose between observations and refine the predicted boxes. This coarse-to-fine design substantially improves localization accuracy, while extensive experiments further examine the effects of visual backbone, image resolution, depth information, viewpoint variation, and synthetic-to-real transfer.

In summary, our contributions are:
\begin{itemize}
\item We formulate retail shelf monitoring as an object-level 3D change detection task, where temporal shelf changes are directly localized as 3D bounding boxes from paired RGB-D observations.
\item We introduce ShelfChange3D, a benchmark consisting of a large-scale synthetic dataset and a real-world dataset, and propose ChangeBox, an end-to-end framework for jointly reasoning across observations and predicting product-level 3D change boxes.
\item We identify precise 3D localization as the remaining bottleneck and introduce a geometry-based refinement stage that significantly improves box accuracy. Extensive experiments demonstrate robustness to viewpoint variation and effective synthetic-to-real transfer.
\end{itemize}

\section{Related Work}
\label{sec:related}

\noindent\textbf{Out-of-Stock Detection.}
Early out-of-stock systems relied on sales data, shelf sensors, or depth-based occupancy estimation~\cite{papakiriakopoulos2009decision,frontoni2018design,milella2020towards}. More recent vision-based approaches detect empty regions using attention maps, segmentation, or object detectors~\cite{allegra2021exploiting,santra2022graph,de2023autonomous,vsikic2024enhanced}. Existing retail datasets mainly provide 2D annotations: SKU-110K~\cite{goldman2019precise} labels visible products with 2D boxes, while EgoCart-OOS~\cite{allegra2021exploiting} and several small-scale datasets~\cite{out-of-stock-b2irt_dataset,empty-slots-in-shelves_dataset,out-of-stock-detection_dataset} annotate missing products or vacant regions in image space. To the best of our knowledge, no existing retail benchmark provides product-level OOS annotations as ground-truth 3D bounding boxes.

\noindent\textbf{Scene Change Detection.}
Scene change detection typically compares observations acquired at different times and predicts changed regions in image space~\cite{alcantarilla2018street,Sakurada2018WeaklySS,chen2021dr,lin2025robust,wang2023reduce,wang2021transcd,Sakurada2017DenseOF,Wu2025SceneDiffAB,jst2015change}. Benchmarks such as VL-CMU-CD~\cite{alcantarilla2018street}, PSCD~\cite{Sakurada2018WeaklySS}, and ChangeSim~\cite{park2021changesim} therefore mainly provide 2D change masks. Recent work extends change detection to 3D through geometric consistency,
segmentation-guided 3D reasoning, and multi-view aggregation
~\cite{adam2022objects,adam2023has,zhou20263d}. 3RScan~\cite{wald2019rio} supports object re-localization across rescans, while DenseChangeCap~\cite{Qiu20233DCL} detects added, removed, and moved objects as 3D boxes from registered point-cloud scans. In contrast, we consider direct object-level 3D change detection from paired RGB-D observations.

\noindent\textbf{3D Object Detection.}
3D object detection has been extensively studied using point clouds~\cite{qi2019deep,zhang2020h3dnet,rukhovich2022fcaf3d,Misra2021AnET} and images~\cite{brazil2023omni3d,Xie2023PixelAlignedRQ,wang2022detr3d,liu2022petr}. Recent methods further explore open-vocabulary and foundation-model-based 3D detection~\cite{yao2026open,lazarow2025cubify,zhang2025detect}. These methods generally assume that the target object is visible in the input observation~\cite{Xu2023RethinkingTL}. For shelf change detection, however, a removed product is absent from the current frame, requiring cross-time reasoning to recover its 3D location.
\section{ShelfChange3D Benchmark}
\subsection{Synthetic Dataset}
\label{sec:dataset}

Collecting and annotating before/after RGB-D pairs of changing shelves in real stores is expensive: each change would require manual staging and 3D box annotation. We therefore build our benchmark in simulation, where the exact pose of every product is known and the same shelf state can be rendered from many viewpoints. The acquisition logic \emph{emulates} a robotic inspection scenario, in which a robot revisits the same shelves at intervals and observes each shelf from a few stopping points with imperfect pose repeatability. 

\subsubsection{\textbf{Problem Definition}}
Given a reference observation ${O}_0$ at time $t_0$ and a current observation ${O}_1$ at time $t_1$ of the same shelf, each consisting of RGB-D with known camera intrinsics and gravity direction, the goal is to predict a set of changes ${C}=\{(B_i, s_i)\}$, where $B_i$ is an oriented 3D box in the camera frame of ${O}_1$ and $s_i\in\{\text{removed},\text{added}\}$ is the change type. We consider the setting where the relative camera pose between the two observations is not externally provided. An added product is directly observable in ${O}_1$, whereas a removed product is visible only in ${O}_0$ and must be localized in the current camera frame.

\begin{figure}[t]
  \centering
  \includegraphics[width=\linewidth]{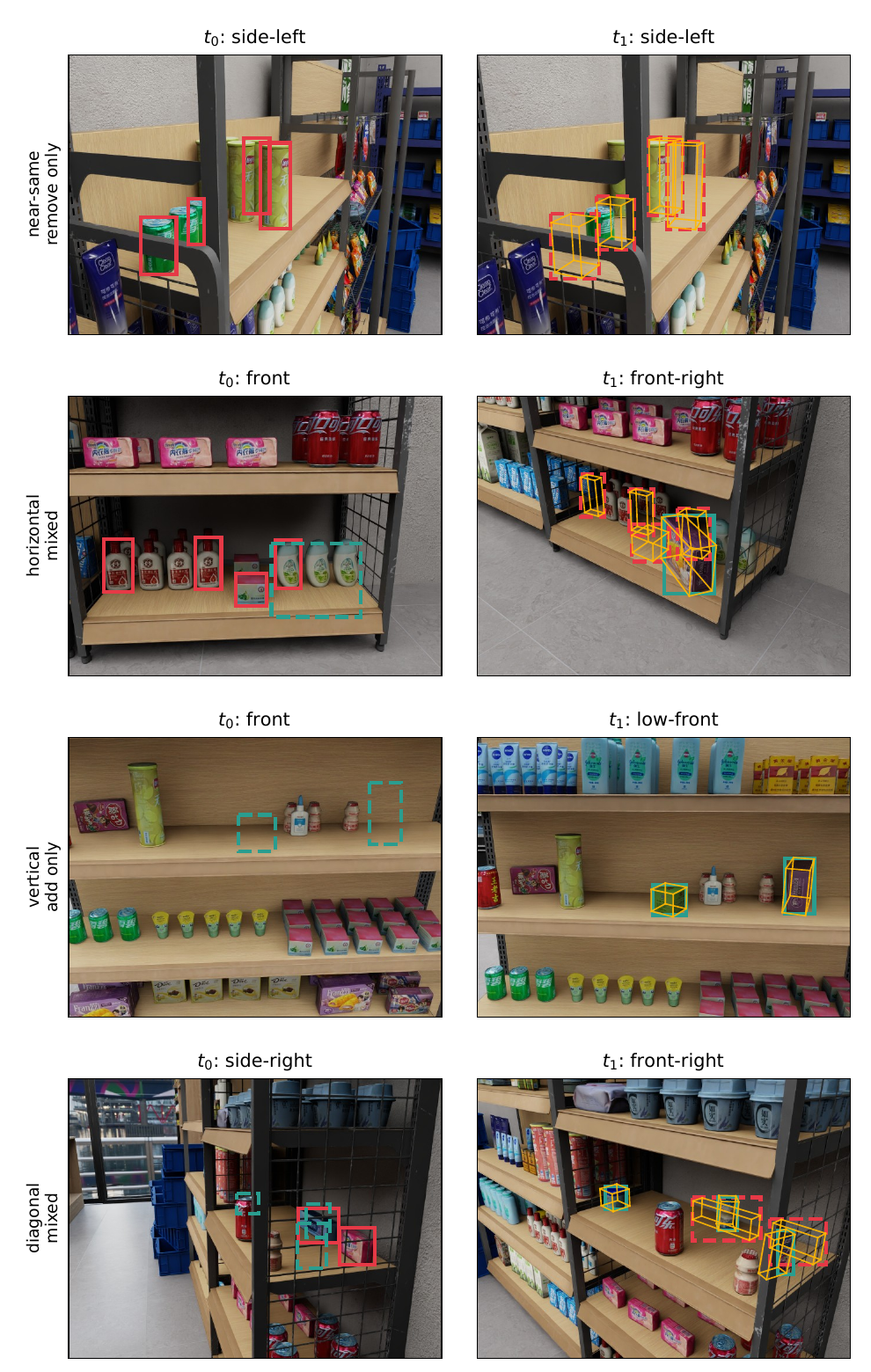}
  \caption{An example of ShelfChange3D-145K, one per viewpoint-change type (rows). Red = removed, green = added.}
  \vskip -0.1in
  \label{fig:examples}
\end{figure}

\subsubsection{\textbf{Generation Process}}
\label{sec:generation}

\begin{figure}[h]
  \centering
  \includegraphics[width=\linewidth]{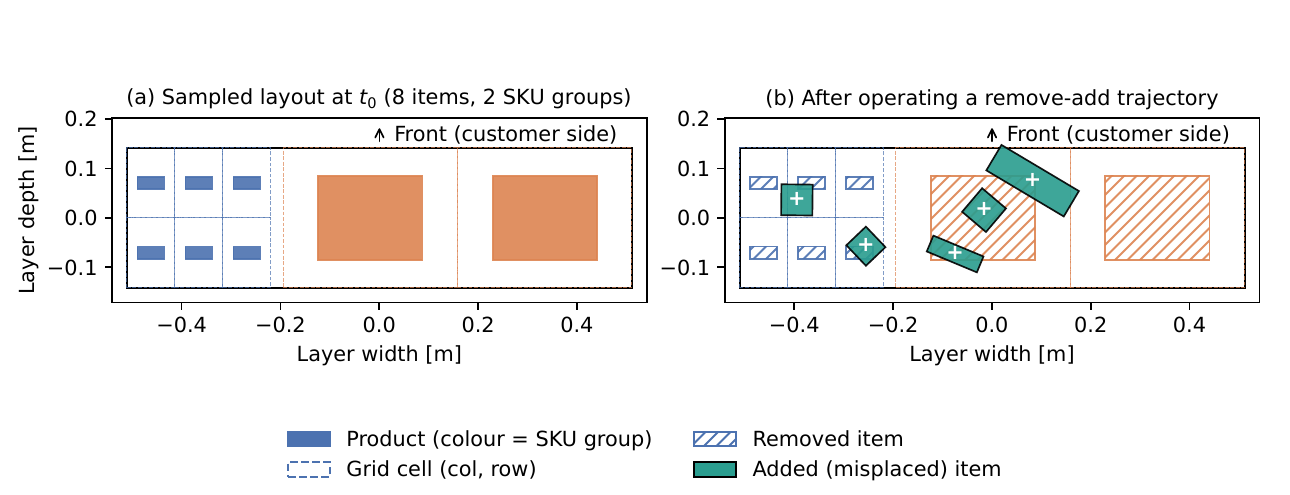}
  \caption{An example of a top-down view of one shelf layer. (a) The sampled layout: two SKU groups with their column/row grid cells. (b) The same layer after remove-add trajectory steps: hatched cells were vacated by removals, green rectangles are misplaced products.}
  \label{fig:layout}
\end{figure}

\begin{figure}[h]
  \centering
  \includegraphics[width=\linewidth]{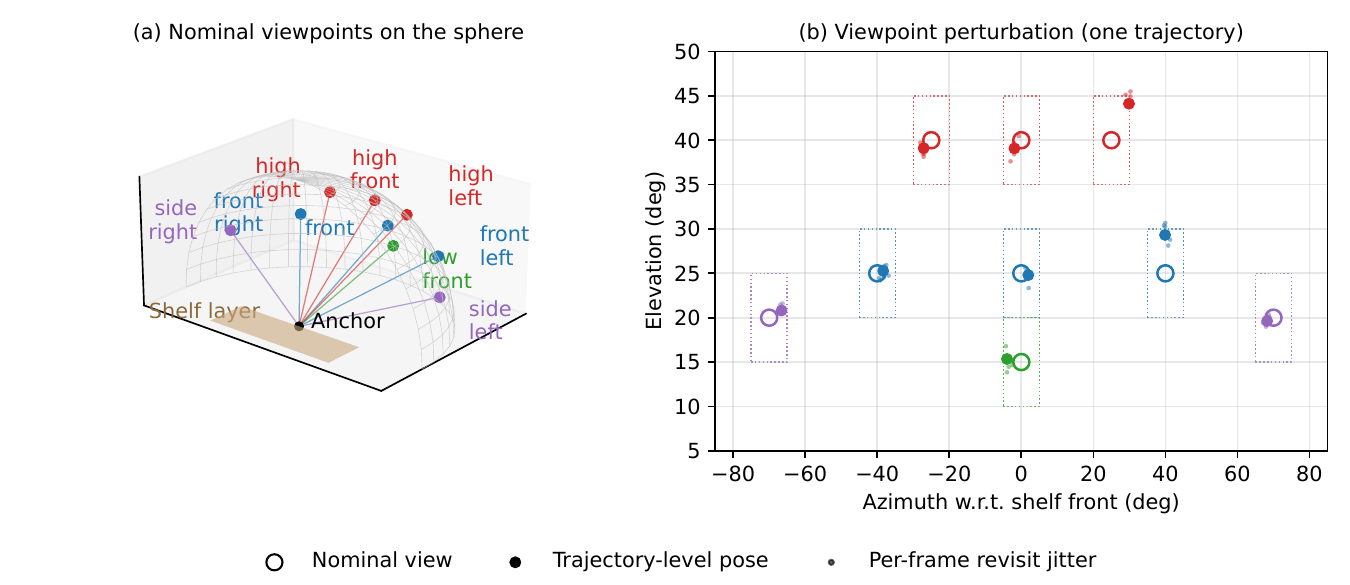}
  \caption{Camera configuration. (a) The nine nominal viewpoints on the viewing sphere around a layer anchor. (b) Azimuth/elevation of the viewpoints for one trajectory: hollow circles are nominal poses, dotted boxes the trajectory-level perturbation range, filled circles the realized trajectory-level pose, and small dots the per-frame revisit jitter over the states of the trajectory.}
  \label{fig:cameras}
\end{figure}

The pipeline is implemented in NVIDIA Isaac Sim~\cite{NVIDIA_Isaac_Sim} using the Genie Sim supermarket scene~\cite{yin2026geniesim30}. \emph{Assets:} The shelf units of the scene are split into individual boards with a known placeable area using Blender~\cite{blender_5_1_1}. \emph{Layouts:} A sampler draws SKU groups with numbers of columns and rows, packs them onto the board, adds small position and yaw noise. An example is shown in Fig.~\ref{fig:layout}. \emph{Cameras:} Cameras emulate an Intel RealSense D435i and are placed on the sphere around each layer at nine nominal viewpoints. A trajectory-level offset, drawn once per camera group, makes every trajectory's initial poses unique, and a per-frame jitter drawn at every render models the pose error of a robot returning to the same stopping point. An example is shown in Fig.~\ref{fig:cameras}. \emph{Changes:} Starting from a fully stocked state $S_0$, a trajectory of states $S_0,\dots,S_n$ is simulated by atomic changes and rendered from all cameras: \emph{removals} take the front-most product of a few columns as customers do, and \emph{additions} place a misplaced SKU from outside the layer's SKU assortment into free space with a random yaw. We generate both remove-only and remove-add trajectories, where each transition between shelf states involves the removal and/or addition of one or more products. 

\subsubsection{\textbf{Pair construction}}
\label{sec:annotation}
Observation pairs are formed offline from the recorded trajectories. Let $S_0,\dots,S_n$ denote the states of one trajectory, ${P}(S)$ the products present in state $S$, ${V}_g$ the set of camera views in group $g$, and $r(S,v)$ the RGB-D frame of state $S$ rendered from view $v$. A sample pairs a reference state $S_i$ observed from view $v_0$ with a later state $S_j$ observed from view $v_1$ in the same group,
\begin{equation}
    \begin{aligned}
    ({O}_0,{O}_1)
    &= \big(r(S_i,v_0),r(S_j,v_1)\big), \quad 
    0\le i<j\le n, \\
    j - i \le \Delta
    &\quad v_0,v_1\in{V}_g .
    \end{aligned}
    \label{eq:syn_pair}
\end{equation}
where $\Delta$ bounds the number of changes between the two observations. The label of the pair lists every product that was removed or added between the two states, together with its 3D box,
\begin{equation}
  \begin{aligned}
  {C}_{ij} = {}&\big\{(B(p),\text{removed}) : p\in{P}(S_i)\setminus{P}(S_j)\big\}\\
  \cup{}&\big\{(B(p),\text{added}) : p\in{P}(S_j)\setminus{P}(S_i)\big\}.
  \end{aligned}
  \label{eq:syn_label}
\end{equation}
Here ${P}(S_i)\setminus{P}(S_j)$ are the products that are on the shelf at $t_0$ but no longer at $t_1$, ${P}(S_j)\setminus{P}(S_i)$ those that are on the shelf at $t_1$ but were not at $t_0$, and products present in both states are not labelled. $B(p)$ is the 3D box of product $p$ where it is transformed into the $t_1$ camera frame. 
Viewpoint pairs are categorized by their nominal cameras: near-same (same camera), horizontal (two cameras on the same elevation ring), vertical (two cameras on the same azimuth ring), and diagonal (all other pairs).

\subsection{Real-World Dataset}
\label{sec:real_dataset}

To measure the gap between simulation and a real store, we additionally collected an OOS dataset, ShelfChange3D-5K, on real-world shelves with an Intel RealSense D435i. Products were removed from the shelf one at a time and every removed product was annotated with a 3D box in the same format as the synthetic data. Since annotating boxes by hand is expensive, we built a semi-automatic pipeline (Fig.~\ref{fig:real_stages}) that turns each removal into a 3D box using image differencing, promptable segmentation~\cite{Carion2025SAM3S}, 3D reconstruction~\cite{Team2025SAM33} with manual review. For the real dataset, we focus on a controlled subset of the task in which the camera remains fixed within each trajectory and only product removals are considered.

\subsubsection{\textbf{Data Acquisition}}
Data were recorded in a laboratory mini-market stocked with retail products. The D435i streams colour and depth, with depth aligned to the colour frame. The gravity direction is stored in the camera frame. 

\subsubsection{\textbf{3D box Annotation Pipeline}}
\label{sec:real_annotation}
\begin{figure}[t]
  \centering
  \includegraphics[width=\linewidth]{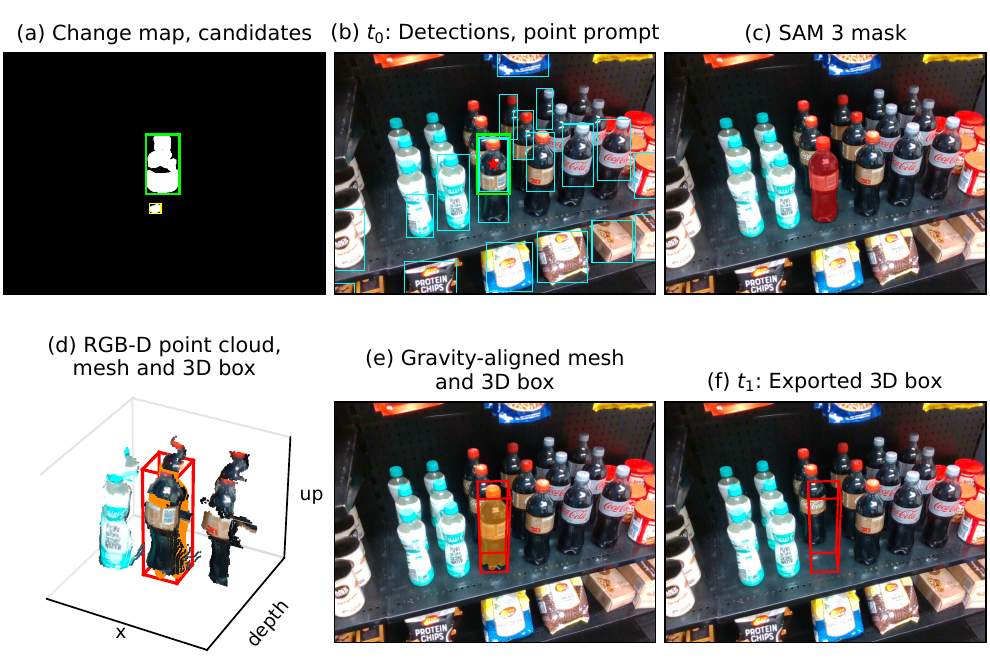}
  \caption{Example of annotation pipeline for one edge. (a)~Change map with difference candidates (yellow) and the one selected by the product detector (green). (b) The selected candidate and the point prompt derived from its centre (red star). (c)~The SAM~3 mask. (d)~The $t_0$ RGB-D point cloud with the SAM~3D Objects mesh placed in it and the box fitted to the mesh. (e)~Translucent mesh and box over $t_0$. (f)~The 3D box over $t_1$, where it marks the vacated slot.}
  \label{fig:real_stages}
\end{figure}

The annotation unit is an \emph{edge} $(S_k,S_{k+1})$ between consecutive frames, which contains one removed product. Fig.~\ref{fig:real_stages} follows one edge through the pipeline. The removed product shows up as a region in the grayscale difference of the two frames. False regions are rejected by keeping the candidate that best overlaps a detection of a YOLO detector~\cite{yolov8_ultralytics} trained on SKU-110K~\cite{goldman2019precise}. The centre of the selected region produces a point prompt for SAM~3~\cite{Carion2025SAM3S}. The masked product is then reconstructed with SAM~3D Objects~\cite{Team2025SAM33}, which outputs a mesh with its pose in the camera frame. The 3D box is the tight oriented bounding box of the mesh. We discovered during the annotation process that the generated meshes are sometimes tilted by a few degrees, so we correct this by rotating the mesh and its 3D box so that the object's approximately vertical local axis is aligned with gravity.

\begin{figure*}[t]
  \centering
  \includegraphics[width=\linewidth]{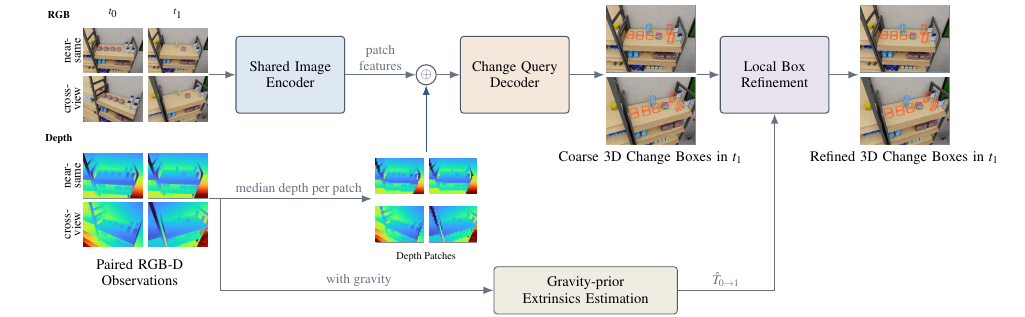}
  \caption{Overview of the proposed method. A shared image encoder extracts patch features from the paired RGB images, while the depths are divided into patches, with each patch represented by its median depth and concatenated to the corresponding RGB feature. The resulting RGB-D tokens are fed into a change-query decoder to predict coarse 3D change boxes, which are further refined by a local box refinement module using the relative extrinsics estimated by the gravity-prior extrinsics module. Orange boxes mark removed products, blue boxes added ones.}
  \label{fig:method}
\end{figure*}

\subsubsection{\textbf{Pair Construction}}
Pairs are built with the same rule as in the synthetic data, Eqs.~\eqref{eq:syn_pair} and \eqref{eq:syn_label}, specialized to a static camera and removals only. The box $B(p)$ of a product removed at edge $(S_k,S_{k+1})$ is the one annotated on Sec.~\ref{sec:real_annotation}. The label of a pair $(S_i,S_j)$ is the set of boxes of the steps $i\le k<j$.

\section{Method}
\label{sec:method}
To address this problem, we develop ChangeBox, an end-to-end framework that detects object-level changes as 3D bounding boxes. Fig.~\ref{fig:method} illustrates the overview of the proposed method. A shared image backbone encodes both observations, with depth appended to the image tokens as an additional dimension, and a query-based decoder predicts 3D boxes for changed objects (Sec.~\ref{sec:detector}). In parallel, the relative camera pose is estimated from the depths using gravity-prior estimation (Sec.~\ref{sec:registration}). Finally, a lightweight network refines each prediction, with points from $O_0$ transformed into the $t_1$ frame for removed objects (Sec.~\ref{sec:sgr}).

\subsection{ChangeBox Detector}
\label{sec:detector}

\subsubsection{\textbf{Input Data}}
The input is paired RGB-D observations of the same shelf, before and after the change. The relative pose $T_{0\to1}$ is unknown, so the network has to establish which part of $O_1$ corresponds to which part of $O_0$.

\subsubsection{\textbf{Image Feature Extraction}}
The same backbone encodes each image into $H_F\times W_F$ tokens of $C$ dimensions,
$F^{2d}_i\in\mathbb{R}^{C\times H_F\times W_F}$. Sharing the backbone puts the tokens of the two frames into one feature space, so that the same product looks the same to the decoder in both frames. 

\subsubsection{\textbf{Depth Dimension}}
For every cell of the feature grid we take the median of the depth values inside it, giving
$\bar{D}_i\in\mathbb{R}^{1\times H_F\times W_F}$ on the feature grid, and concatenate it with features,
\begin{equation}
  F_i = \phi\big([F^{2d}_i;\ \bar{D}_i]\big), \qquad i = 0, 1,
  \label{eq:depth}
\end{equation}
where $[\cdot;\cdot]$ concatenates along the dimension and $\phi$ is a small convolutional projection. Every token now carries the appearance of its patch and its metric depth.

\subsubsection{\textbf{Change Query Decoder}}
The tokens of both frames are joined into one sequence of $2\,H_F W_F$ tokens. We follow the standard transformer decoder in DETR~\cite{Carion2020EndtoEndOD}, which includes L decoder layers. Following PETR~\cite{liu2022petr}, queries are initialized from learnable 3D anchor points and updated by a transformer decoder,
\begin{equation}
  Q_l = \Omega_l\big([F_0; F_1],\ Q_{l-1}\big), \qquad l = 1, \dots, L,
  \label{eq:decoder}
\end{equation}
where $\Omega_l$ is a decoder layer. The output $Q_L\in\mathbb{R}^{K\times C}$ holds one region feature per candidate change. 

\subsubsection{\textbf{Detection Head}}
From each region feature, the head predicts $\hat{y}=(\hat{\mathbf{s}}, \hat{\mathbf{b}})$:
the probability distribution $\hat{\mathbf{s}}$ over the change types \{removed, added\} and
the box $\hat{\mathbf{b}}=[\hat{\mathbf{c}}, \hat{\mathbf{d}}, \hat\theta]$ with center
$\hat{\mathbf{c}}\in\mathbb{R}^3$, size $\hat{\mathbf{d}}\in\mathbb{R}^3$ and yaw $\hat\theta$ about the vertical axis. All boxes are expressed in the gravity-aligned $t_1$ camera frame.

\subsection{Gravity-prior Extrinsics Estimation}
\label{sec:registration}

The detector needs no extrinsics, but the refinement stage does. We estimate the relative pose $\hat{T}$ from the two depth maps, mapping $t_0$ to $t_1$. The gravity direction is known in both frames, making this estimation possible. The estimator has five stages.

\subsubsection{\textbf{Gravity Alignment}}
Each depth map is back-projected into a point cloud and rotated so that the gravity direction becomes its vertical axis. The two clouds then differ only by a rotation $\theta$ about the vertical axis and a translation $\mathbf{t}\in\mathbb{R}^3$. Because only one angle is left, we can search it exhaustively.

\subsubsection{\textbf{Identity Short-cut}}
Most revisits are taken from nearly the same viewpoint. We first initialize the relative pose with the identity transform and refine it using point-to-plane ICP~\cite{Chen1991ObjectMB}, then evaluate the result based on depth agreement
\begin{equation}
  a(T) = \frac{1}{|{P}_0|}\sum_{\mathbf{p}\in{P}_0}
  \mathbf{1}\big[\,|z(T\mathbf{p}) - D_1(\pi_1(T\mathbf{p}))| < \tau_a\big],
  \label{eq:agreement}
\end{equation}
where ${P}_0$ is the point cloud of $t_0$, $T\mathbf{p}$ is a point moved into the
$t_1$ frame, $z(\cdot)$ is its depth, $\pi_1(\cdot)$ projects it to a pixel of $t_1$,
$D_1(\cdot)$ is the depth measured at that pixel, $\mathbf{1}[\cdot]$ is the indicator
function and $\tau_a$ a distance tolerance. If the agreement is high, the identity solution goes directly to the forward-backward check.

\subsubsection{\textbf{Candidate Generation}}
Otherwise $\theta$ is swept from $-100^\circ$ to $+100^\circ$ in steps of $3^\circ$. For each candidate angle, we project both point clouds onto top-down occupancy grids and estimate the horizontal translation using FFT-based cross-correlation. The vertical translation is found in the same way from the height histograms of the two clouds. We retain several high-scoring translation hypotheses for each angle. The final candidate set consists of these top hypotheses across angles, together with the identity transform.

\subsubsection{\textbf{Refinement and Selection}}
Every remaining candidate is refined by a few iterations of point-to-plane ICP and ranked by the agreement in Eq.~\eqref{eq:agreement}. Only the best few are refined further and re-ranked by the symmetric agreement $\min\big(a(T),\,a'(T^{-1})\big)$, where $a'$ is Eq.~\eqref{eq:agreement} with the roles of the two frames exchanged, i.e.\ the $t_1$ points are moved into $t_0$ by $T^{-1}$ and compared with $D_0$. A wrong solution can occasionally look good in one direction but rarely in both. The candidate with the highest symmetric agreement becomes $\hat{T}$.

\subsubsection{\textbf{Forward-backward Check}}
Finally, $\hat{T}$ is checked by the forward-backward test. We run the point-to-plane ICP once more in the backward direction ($t_1\to t_0$), with $\hat{T}^{-1}$ as the initial guess, which gives an independent backward solution $\hat{T}'$. If $\hat{T}$ is correct, the backward ICP barely moves and $\hat{T}\hat{T}'\approx I$. The gate compares $\hat{T}\hat{T}'$ with the identity and accepts $\hat{T}$ only if the rotation and translation of this residual transform are below the thresholds $\tau_R$ and $\tau_t$. Otherwise, the pair is treated as unregistered, and refinement is skipped for removed boxes.

\subsection{Local Box Refinement}
\label{sec:sgr}

\begin{table}[h]
    \small
  \centering
  \caption{Error analysis of ChangeBox and evaluation of the refinement on ShelfChange3D-145K.}
  \label{tab:error_analysis}
  \footnotesize
  \setlength{\tabcolsep}{6pt}
  \begin{tabular}{lrrr}
    \toprule
    Method & mAP$_{0.25}$ \up & mAP$_{0.5}$ \up & mAP$_{0.7}$ \up \\
    \midrule
    ChangeBox & 87.0 & 57.5 & 14.8 \\
    Size $\rightarrow$ GT & 87.2 & 57.7 & 14.8 \\
    Center $\rightarrow$ GT & 97.1 & 95.6 & 94.0 \\
    Both $\rightarrow$ GT & \textbf{97.4} & \textbf{96.7} & \textbf{95.5} \\
    \midrule
    ChangeBox+Refine & 93.0 & 81.7 & 48.3 \\
    \bottomrule
  \end{tabular}
\end{table}

Our ChangeBox detector localizes products to about 1 \, cm, but its boxes rarely reach an IoU of 0.7. Tab.~\ref{tab:error_analysis} shows why: replacing the predicted centers by the ground truth raises mAP$_{0.7}$ from 14.8 to 94.0, while replacing the sizes has little change. The boxes are shifted, not mis-sized. We therefore refine each predicted box using the depth points enclosed by the box.

\subsubsection{\textbf{Point Collection}}
An \emph{added} box uses the $t_1$ depth, with a \emph{removed} box uses the $t_0$ depth moved into the $t_1$ frame with $\hat{T}_{0\to1}$. Following PointRCNN~\cite{Shi2018PointRCNN3O}, we slightly enlarge each predicted box, collect the enclosed points, and transform them into the box coordinate frame. Each point is then represented by its normalized position, signed distances to the six box faces, depth normal, depth-gradient magnitude, local valid-depth ratio, and depth, together with the predicted box score, class, and log size.

\subsubsection{\textbf{Refinement Head}}
Local points are encoded by a shared point-wise MLP and aggregated with max and mean pooling. A regression head then predicts a center correction $\Delta\mathbf{c}$ and a log-scale size correction $\Delta\log\mathbf{d}$. The refined center and size are $\mathbf{c}+R\Delta\mathbf{c}$ and $\mathbf{d}\odot e^{\Delta\log\mathbf{d}}$, while the orientation remains unchanged.

\subsection{Loss Function}
\label{sec:loss}
\subsubsection{\textbf{3D change detection Loss}}
Let $y=(\mathbf{s}, \mathbf{b})$ and $\hat{y}=(\hat{\mathbf{s}}, \hat{\mathbf{b}})$ denote the
ground-truth and predicted change boxes of a pair, where $\mathbf{s}$ is the change type and
$\mathbf{b}=[\mathbf{c}, \mathbf{d}, \theta]$ the box. The Hungarian algorithm is used for label assignment between ground-truths and predictions. With $\sigma$ the optimal assignment, the detection loss is
\begin{equation}
  {L}(y, \hat{y}) = \lambda_{cls}\,{L}_{cls}\big(\mathbf{s}, \sigma(\hat{\mathbf{s}})\big)
  + {L}_{reg}\big(\mathbf{b}, \sigma(\hat{\mathbf{b}})\big),
  \label{eq:loss}
\end{equation}
where ${L}_{cls}$ is the focal loss~\cite{Lin2017FocalLF} over the change types and
\begin{equation}
  \begin{split}
  {L}_{reg} = \lambda_c\,\|\hat{\mathbf{c}}-\mathbf{c}\|_1
  &+ \lambda_d\,\|\log\hat{\mathbf{d}}-\log\mathbf{d}\|_1 \\
  &+ \lambda_\theta\,\min_{k}\big\|\hat{\mathbf{r}}-\mathbf{r}(\theta + 2\pi k/m)\big\|_1 ,
  \end{split}
  \label{eq:loss_reg}
\end{equation}
with $\mathbf{r}(\theta)=[\sin\theta, \cos\theta]$. The orientation term is taken against the
closest of the $m$ equivalent yaw angles of a product with symmetry order $m$. 

\subsubsection{\textbf{Refinement Loss}}
The refinement network is trained separately with
\begin{equation}
  {L}_{ref} = \|\Delta\mathbf{c}-\Delta\mathbf{c}^\ast\|_1
  + \|\Delta\mathbf{d}-\Delta\mathbf{d}^\ast\|_1 ,
  \label{eq:loss_ref}
\end{equation}
where $\Delta\mathbf{c}^\ast$ and $\Delta\mathbf{d}^\ast$ map the input box to its ground truth.

\section{Experiments}
\label{sec:experiments}

\subsection{Experimental Setup}
\label{sec:exp_setup}

\noindent\textbf{Evaluation Metrics.}
We evaluate change detection using 3D detection evaluation metrics mAP and mAR at IoU thresholds of 0.25, 0.5, and 0.7. Results are computed separately for removed and added objects and then averaged across the two change types. On the real dataset, which contains only removals, metrics are computed over the removed class. To measure the quality of the boxes independently of the ranking we report the mean center error (c-err) in centimeters. During evaluation, we retain at most 20 predicted detections for each observation pair.

\noindent\textbf{Baselines.}
We consider the DenseChangeCap~\cite{Qiu20233DCL} as baseline in comparison that localize 3D changes from point clouds: it takes the point clouds of the two observations, differences them, and predicts 3D boxes of the changed objects with a decoder. To apply it to our task, the two depth maps are back-projected into point clouds in $t_1$ camera frames, and the before / after / before$\setminus$after / after$\setminus$before clouds of the DenseChangeCap are built from them. 

\noindent\textbf{Implementation Details.}
The image backbone is pluggable, and we compare three similarly sized models: an ImageNet-pretrained ResNet-50, and the ConvNeXt-Tiny and DINOv3-S models distilled from DINOv3~\cite{Simoni2025DINOv3}. DINOv3-S is used by default unless stated otherwise. The refinement network is trained once on DINOv3-S predictions and reused for all backbones and real-data experiments. ShelfChange3D-145K is split into training, validation, and test sets at 0.7:0.15:0.15, while ShelfChange3D-5K contains 5{,}520 pairs, with 3{,}210 for training and 2{,}310 for testing.

\begin{figure*}[!t]
  \centering
  \includegraphics[width=\linewidth]{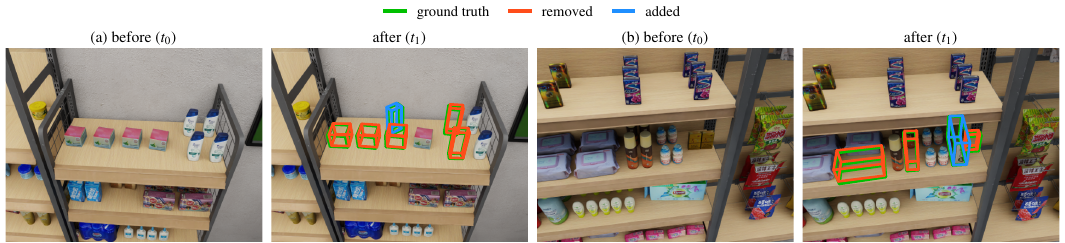}
  \caption{Qualitative results. Example results for ChangeBox on ShelfChange3D-145K.}
      \vskip -0.07in
  \label{fig:qual}
\end{figure*}

\subsection{Experiments on the ShelfChange3D-145K}

\label{sec:exp_syn}

\noindent\textbf{Base Experiments.}
As shown in Tab.~\ref{tab:main}, ChangeBox substantially outperforms the DenseChangeCap baseline, with DINOv3-S achieving much higher mAP across all IoU thresholds. The refinement stage improves all three backbones, with particularly large gains at stricter IoU thresholds and lower center error, confirming its effectiveness for precise 3D localization. Among all configurations, DINOv3-S with refinement performs best, reaching 93.0, 81.7, and 48.3 mAP at IoU thresholds of 0.25, 0.5, and 0.7, respectively.

\begin{table}[!t]
    \small
  \centering
  \caption{Comparison of ChangeBox variants with the baseline on ShelfChange3D-145K.}
  \label{tab:main}
  \footnotesize
  \setlength{\tabcolsep}{2pt}
  \begin{tabular}{lcccccccc}
    \toprule
    \multirow{2}{*}{Method}
    & \multicolumn{3}{c}{mAP \up}
    & \multicolumn{3}{c}{mAR \up}
    & \multicolumn{2}{c}{c-err (cm) \down} \\
    \cmidrule(lr){2-4} \cmidrule(lr){5-7} \cmidrule(lr){8-9}
     & 0.25 & 0.5 & 0.7 & 0.25 & 0.5 & 0.7 & 0.25 & 0.5 \\
    \midrule
    DenseChangeCap~\cite{Qiu20233DCL}
      & 39.1 & 4.1 & 0.0 & 66.0 & 20.7 & 1.7 & 1.96 & 1.19 \\
    \midrule
    ChangeBox (R)
      & 79.1 & 46.2 & 10.0 & 85.1 & 61.0 & 26.4 & 1.29 & 0.98 \\
    ChangeBox (R) + refine
      & 88.6 & 74.2 & 36.0 & 92.3 & 81.5 & 53.2 & 0.89 & 0.73 \\
    \midrule
    ChangeBox (C)
      & 85.9 & 57.3 & 16.3 & 89.5 & 69.2 & 33.3 & 1.14 & 0.91 \\
    ChangeBox (C) + refine
      & 92.2 & 80.6 & 45.4 & 94.5 & 85.8 & 60.5 & 0.78 & 0.66 \\
    \midrule
    ChangeBox (D)
      & 87.0 & 57.5 & 14.8 & 90.5 & 69.9 & 32.8 & 1.14 & 0.92 \\
    ChangeBox (D) + refine
      & \textbf{93.0} & \textbf{81.7} & \textbf{48.3}
      & \textbf{95.2} & \textbf{86.7} & \textbf{62.9}
      & \textbf{0.75} & \textbf{0.64} \\
    \bottomrule
    \multicolumn{9}{l}{\scriptsize R: ResNet-50; C: ConvNeXt-T; D: DINOv3-S.}
  \end{tabular}
\end{table}

\begin{table}[!t]
    \small
  \centering
  \caption{Evaluation of different visual representation and resolution on ShelfChange3D-145K.}
  \label{tab:abl_backbone}
  \footnotesize
  \setlength{\tabcolsep}{4pt}
  \begin{tabular}{lcrrrr}
    \toprule
    Method & Res. & mAP$_{0.25}$ \up & mAP$_{0.5}$ \up & mAR$_{0.25}$ \up & c-err$_{0.25}$ \down \\
    \midrule
    ChangeBox (R) & 320 & 61.9 & 23.9 & 72.1 & 1.67 \\
    ChangeBox (C) & 320 & 75.0 & 37.3 & 82.5 & 1.44 \\
    ChangeBox (D) & 320 & 73.6 & 35.8 & 82.0 & 1.44 \\
    \midrule
    ChangeBox (R) & 640 & 79.1 & 46.2 & 85.1 & 1.29 \\
    ChangeBox (C) & 640 & 85.9 & 57.3 & 89.5 & 1.14 \\
    ChangeBox (D) & 640 & \textbf{87.0} & \textbf{57.5} & \textbf{90.5} & \textbf{1.14} \\
    \bottomrule
    \multicolumn{6}{l}{\scriptsize R: ResNet-50; C: ConvNeXt-T; D: DINOv3-S.} \\
    \multicolumn{6}{l}{\scriptsize Res: 320 = $320\times240$; 640 = $640\times480$.}
\end{tabular}
    \vskip -0.12in
\end{table}

\noindent\textbf{Visual Representation and Resolution.}
As shown in Tab.~\ref{tab:abl_backbone}, stronger visual backbones and higher input resolution consistently improve performance. At 640 resolution, ConvNeXt-T and DINOv3-S outperform ResNet-50, while increasing the resolution from 320 to 640 further improves all three backbones, particularly at stricter IoU thresholds. DINOv3-S at 640 resolution achieves the best overall performance. These results indicate that richer visual representations and higher spatial resolution are important for accurately localizing small, densely packed products.

\noindent\textbf{Explicit Depth.}
Tab.~\ref{tab:abl_depth} shows the effect of the depth information. Explicit depth information consistently improves performance across all three backbones. Adding depth increases both mAP and mAR, with the largest gains generally observed at the stricter IoU threshold. For DINOv3-S, mAP$_{0.5}$ rises from 45.1 to 57.5 and the center error decreases from 1.28 cm to 1.14 cm. ResNet-50 and ConvNeXt-T show similar improvements. These results indicate that explicit depth provides complementary geometric cues for more accurate 3D localization.

\begin{table}[!t]
    \small
  \centering
  \caption{Evaluation of effect of the depth information on ShelfChange3D-145K.}
  \label{tab:abl_depth}
  \footnotesize
  \setlength{\tabcolsep}{4pt}
  \begin{tabular}{lcrrrr}
    \toprule
    Method & Depth & mAP$_{0.25}$ \up & mAP$_{0.5}$ \up & mAR$_{0.25}$ \up & c-err$_{0.25}$ \down \\
    \midrule
    ChangeBox (R) & \xmark & 72.4 & 38.4 & 81.0 & 1.36 \\
    ChangeBox (R) & \cmark & 79.1 & 46.2 & 85.1 & 1.29 \\
    \midrule
    ChangeBox (C) & \xmark & 83.0 & 52.3 & 88.1 & 1.16 \\
    ChangeBox (C) & \cmark & 85.9 & 57.3 & 89.5 & 1.14 \\
    \midrule
    ChangeBox (D) & \xmark & 80.5 & 45.1 & 86.7 & 1.28 \\
    ChangeBox (D) & \cmark & \textbf{87.0} & \textbf{57.5} & \textbf{90.5} & \textbf{1.14} \\
    \bottomrule
    \multicolumn{6}{l}{\scriptsize R: ResNet-50; C: ConvNeXt-T; D: DINOv3-S.}
  \end{tabular}
      \vskip -0.12in
\end{table}

\begin{figure}[!h]
  \centering
    \vskip -0.07in
  \includegraphics[width=\linewidth]{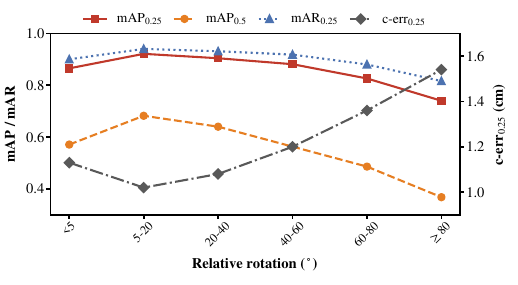}
  \vskip -0.12in
  \caption{Accuracy and center error vs.\ relative camera rotation between $t_0$ and $t_1$. The $<5^\circ$ bin holds 61\% of the test pairs.}
  \label{fig:viewpoint}
\end{figure}
\noindent\textbf{Robustness to Viewpoint Change.}
Fig.~\ref{fig:viewpoint} splits the test set by the relative camera rotation. The model remains robust to moderate viewpoint changes, with strong mAP$_{0.25}$ and mAR$_{0.25}$ up to about $40^\circ$, while performance degrades at larger rotations, especially in mAP$_{0.5}$ and center accuracy. Notably, the $<5^\circ$ range is not the best-performing bin despite containing most training pairs, indicating that performance is not solely determined by training frequency.

\begin{figure}[!t]
  \centering
  \includegraphics[width=\linewidth]{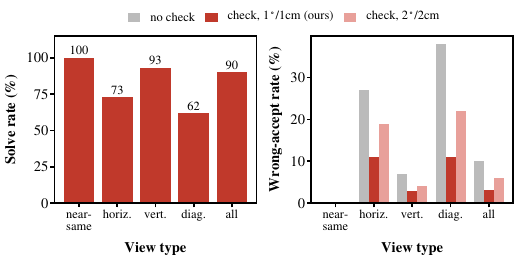}
  \vskip -0.07in
  \caption{Extrinsics estimation on the test set. Left: solve rate per view type. Right: wrongly accepted per view type, both with and without the forward-backward check.}
  \label{fig:registration}
\end{figure}

\noindent\textbf{Extrinsics Estimation.}
Fig.~\ref{fig:registration} evaluates the extrinsics estimator on the test set. It achieves a 90\% overall solve rate, reaching 100\% for near-same views but dropping to 62\% for diagonal pairs. With $\tau_R=1^\circ$ and $\tau_t=1\,\mathrm{cm}$, the forward-backward consistency check reduces the overall wrong-accept rate from about 10\% to 3\%, effectively rejecting unreliable pose estimates while preserving a high solve rate.

\begin{table}[!t]
  \centering
  \caption{Evaluation of Sim2Real transfer on ShelfChange3D-5K.}
  \label{tab:sim2real}
  \footnotesize
  \setlength{\tabcolsep}{2pt}

  \begin{tabular}{cccrrrr}
    \toprule
    \multicolumn{3}{c}{Training} & \multicolumn{4}{c}{Metrics} \\
    \cmidrule(lr){1-3} \cmidrule(lr){4-7}
    Syn.\ pre-train & Real & Refine
    & mAP$_{0.25}$ \up
    & mAP$_{0.5}$ \up
    & mAR$_{0.25}$ \up
    & c-err$_{0.25}$ \down \\
    \midrule

    \xmark & \cmark & \xmark
    & 21.0 & 0.8 & 44.2 & 3.34 \\

    \cmark & \xmark & \xmark
    & 46.7 & 2.2 & 65.3 & 2.87 \\

    \cmark & \cmark & \xmark
    & 92.9 & 37.5 & 95.6 & 2.20 \\

    \cmark & \cmark & \cmark
    & \textbf{97.9} & \textbf{50.1} & \textbf{98.5} & \textbf{1.95} \\
    \bottomrule
  \end{tabular}
  \vskip -0.07in
\end{table}

\noindent\textbf{Qualitative Results.}
Qualitative results are shown in Fig.~\ref{fig:qual}. In both examples, the model correctly identifies multiple changes and distinguishes between removals and additions. The predicted 3D boxes closely overlap the corresponding ground-truth boxes, indicating accurate localization of changed products despite dense shelf arrangements and multiple changes within the same observation pair.

\subsection{Sim2Real on ShelfChange3D-5K}
\label{sec:exp_sim2real}

Synthetic pre-training provides a clear benefit for real-world performance, as shown in Tab.~\ref{tab:sim2real}. Training only on the real-world dataset yields limited performance, whereas the model trained on synthetic data already transfers effectively to the real test set in a zero-shot setting. Fine-tuning the synthetically pre-trained model on real data leads to a substantial further improvement, reaching 92.9 mAP$_{0.25}$ before refinement. The local refinement stage further improves localization quality, increasing mAP$_{0.25}$ to 97.9 while also reducing the center error. Overall, these results demonstrate that the synthetic dataset captures transferable features for real-world 3D change detection and serves as an effective source of pre-training for subsequent real-data adaptation.
\section{Conclusion}
\label{sec:conclusion}

In this paper, we formulate retail shelf monitoring as an object-level 3D change detection problem, where removed and added products are localized as 3D bounding boxes from paired RGB-D observations. To support this setting, we introduce ShelfChange3D, consisting of complementary large-scale synthetic and real-world datasets, together with ChangeBox, an end-to-end framework for jointly reasoning across observations and predicting product-level 3D changes. Our experiments show that ChangeBox effectively detects shelf changes across different viewpoints, while also revealing precise 3D localization as a major remaining challenge. To address this, we introduce a geometry-based refinement stage that leverages depth observations and gravity-prior pose estimation to substantially improve box localization. We further demonstrate effective synthetic-to-real transfer, highlighting the value of ShelfChange3D for both benchmarking and real-world adaptation. Future work will extend the framework to more diverse retail environments and more complex forms of shelf change.






\bibliographystyle{IEEEtran}
\bibliography{refs}

\clearpage
\section*{APPENDIX}

\setcounter{section}{0}
\setcounter{subsection}{0}
\renewcommand{\thesubsection}{\Alph{subsection}}

\subsection{Restocking robot experiments}
\label{app:robot}

\begin{figure*}[h]
  \centering
  \includegraphics[width=\linewidth]{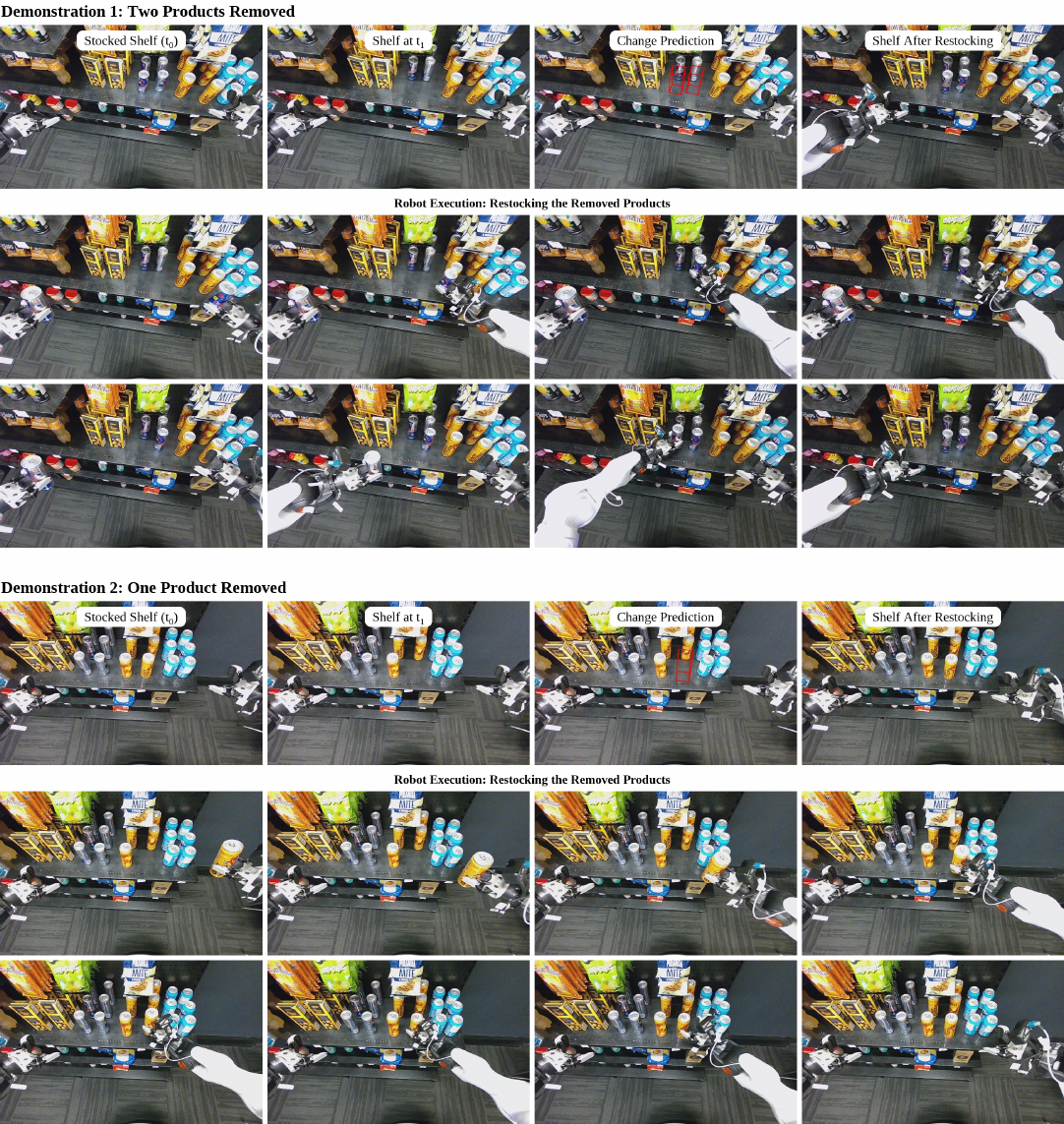}
  \caption{Restocking Robot. For each demonstration, the top row shows the head-camera views: the fully stocked shelf at $t_0$, the shelf at $t_1$ after products were removed, the predicted boxes of the removed products, and the shelf after restocking. Subsequent panels show robots demonstration.}
  \label{fig:robot}
\end{figure*}

In Fig.~\ref{fig:robot}, we show a demonstration of a robot restocking a shelf back to its fully stocked state by applying our method to find the removed products.

\subsubsection{\textbf{Setup}}
We demonstrate our method on a dual-arm humanoid robot. The robot captures a fully stocked shelf at $t_0$ and a current view at $t_1$ after product removal. Our method then localizes each removed product as an oriented 3D box in the $t_1$ camera frame.

\subsubsection{\textbf{Robot Restocking}}
For each detected removed product, we use the center of its predicted 3D box as the placement target. The robot then uses inverse kinematics to place the product at the target, repeating this process for each removed product.

\subsection{Additional experiments}
\subsubsection{\textbf{Training Strategy}}
We compare three training schedules: FT, where the image backbone is fine-tuned throughout training. Frozen, where the backbone remains fixed and only the detection decoder is trained. and Frozen$\rightarrow$FT, where the backbone is initially frozen and subsequently unfrozen for joint fine-tuning. As in Tab.~\ref{tab:abl_train}, the frozen$\rightarrow$FT schedule consistently achieves the best performance across all three backbones. For example, with DINOv3-S, it improves mAP${0.25}$ from 76.2 with FT and 77.0 with a frozen backbone to 87.0, while mAP${0.5}$ increases to 57.5 and the center error decreases to 1.14 cm. Similar gains are observed for ResNet-50 and ConvNeXt-T, suggesting that first training with a fixed visual representation and then jointly adapting the backbone provides more stable and effective optimization.

\begin{table}[h]
  \centering
      \small
  \caption{Evaluation of different training schedules on the ShelfChange3D-145K.}
  \label{tab:abl_train}
  \footnotesize
  \setlength{\tabcolsep}{2pt}

  \begin{threeparttable}
  \begin{tabular}{llrrrr}
    \toprule
    Method & Schedule & mAP$_{0.25}$ \up & mAP$_{0.5}$ \up & mAR$_{0.25}$ \up & c-err$_{0.25}$ \down \\
    \midrule
    ChangeBox (R) & FT & 66.5 & 28.5 & 77.4 & 1.56 \\
    ChangeBox (R) & frozen & 64.1 & 24.7 & 74.3 & 1.66 \\
    ChangeBox (R) & frozen$\to$FT & \textbf{79.1} & \textbf{46.2} & \textbf{85.1} & \textbf{1.29} \\
    \midrule
    ChangeBox (C) & FT\tnote{$\dagger$} & 60.4 & 17.8 & 74.2 & 1.86 \\
    ChangeBox (C) & frozen & 72.9 & 33.7 & 80.8 & 1.51 \\
    ChangeBox (C) & frozen$\to$FT & \textbf{85.9} & \textbf{57.3} & \textbf{89.5} & \textbf{1.14} \\
    \midrule
    ChangeBox (D) & FT & 76.2 & 39.1 & 84.8 & 1.35 \\
    ChangeBox (D) & frozen & 77.0 & 38.8 & 83.5 & 1.45 \\
    ChangeBox (D) & frozen$\to$FT & \textbf{87.0} & \textbf{57.5} & \textbf{90.5} & \textbf{1.14} \\
    \bottomrule
  \end{tabular}

  \begin{tablenotes}[flushleft]
    \footnotesize
    \item[$\dagger$] Training diverged. Resumed from the best checkpoint and continued with the linear decay stage.
  \end{tablenotes}
  {\scriptsize R: ResNet-50; C: ConvNeXt-T; D: DINOv3-S.}
  \end{threeparttable}
\end{table}

\subsubsection{\textbf{Qualitative Results on ShelfChange3D-5K}}
\begin{figure*}[h]
  \centering
  \includegraphics[width=\linewidth]{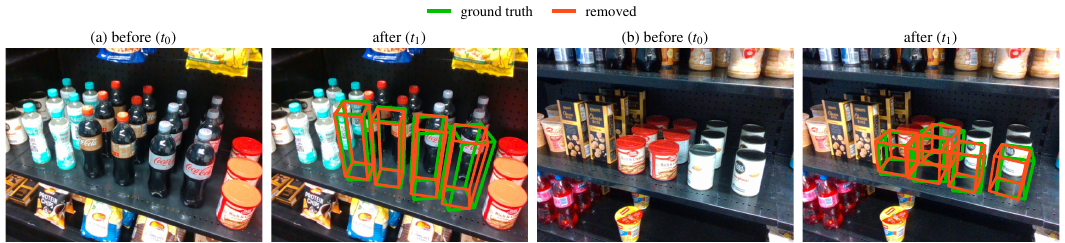}
  \caption{Qualitative results. Example results for ChangeBox on ShelfChange3D-5K.}
      \vskip -0.07in
  \label{fig:qual_real}
\end{figure*}

Fig.~\ref{fig:qual_real} presents qualitative results of ChangeBox on ShelfChange3D-5K. Across both examples, the predicted removed-object boxes closely align with the ground-truth annotations, despite cluttered shelf layouts and partial occlusions. The results illustrate that ChangeBox can accurately localize multiple product removals in real-world shelf scenes.

\begin{figure*}[h]
  \centering
  \includegraphics[width=\linewidth]{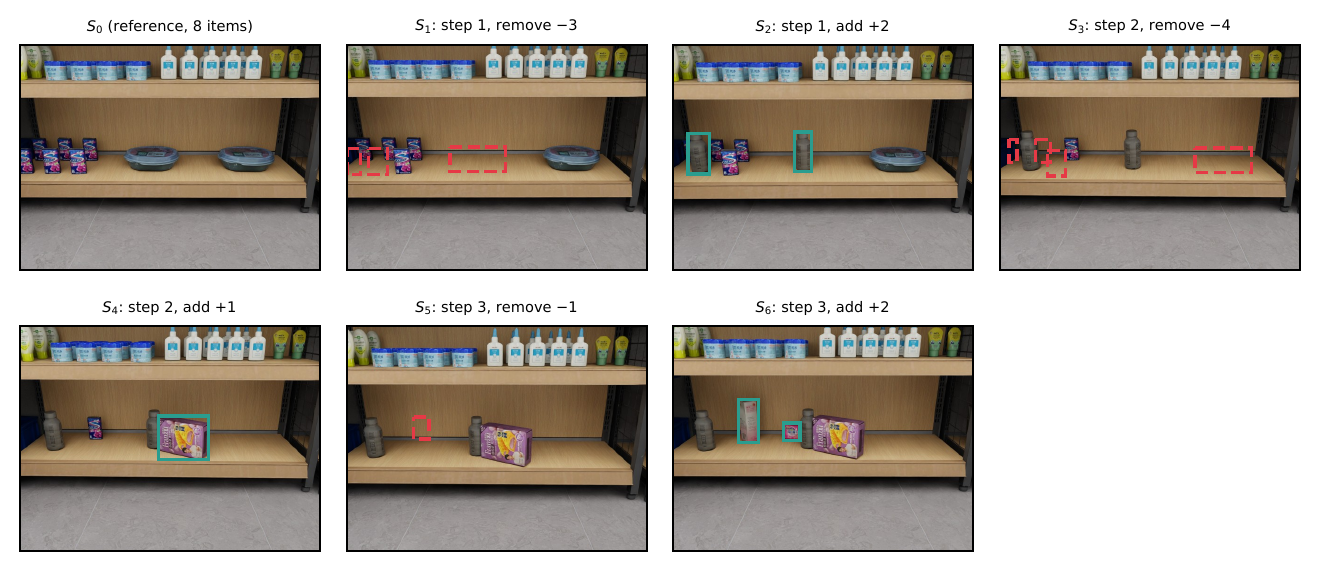}
  \caption{One remove-add trajectory seen from the \texttt{front} camera. Dashed red boxes mark the products removed at that step at their previous position, solid green boxes mark the misplaced products added at that step.}
  \label{fig:trajectory}
\end{figure*}

\subsection{Dataset Novelty}
\label{sec:dataset_novelty}

Tab.~\ref{tab:dataset_comparison} compares ShelfChange3D-145K with existing change
detection and retail datasets by input, task and annotation. \emph{2D change datasets} (VL-CMU-CD~\cite{alcantarilla2018street}, ChangeSim~\cite{park2021changesim}, PSCD~\cite{Sakurada2018WeaklySS}) localize changes in image space only, as masks or 2D boxes. \emph{3D change datasets} label changes at the object level but not as 3D boxes: instance poses in 3RScan~\cite{wald2019rio} and instance masks in SceneDiff~\cite{Wu2025SceneDiffAB}. Only DyS2Change~\cite{Qiu20233DCL} provides 3D boxes, but for pairs of point-cloud scans rather than for single frames. \emph{Retail datasets} (SKU-110K~\cite{goldman2019precise}, EgoCart-OOS~\cite{allegra2021exploiting}) are single images with 2D boxes of visible products or points on shelf gaps, without any before/after pair.

\begin{table}[h]
  \centering
  \caption{Comparison with existing change detection datasets and retail datasets.}
  \label{tab:dataset_comparison}
  \scriptsize
  \setlength{\tabcolsep}{2.5pt}
  \begin{tabular}{lllll}
    \toprule
    Dataset & Domain & Task & Scale & Annotation \\
    \midrule
    \multicolumn{5}{@{}l}{\textit{Change detection in image space}} \\
    VL-CMU-CD~\cite{alcantarilla2018street} & street & segmentation & 1,362 pairs & 2D mask \\
    ChangeSim~\cite{park2021changesim} & industrial & segmentation & 130K images & 2D mask \\
    PSCD~\cite{Sakurada2018WeaklySS} & street & segmentation & 770 pairs & 2D mask \\
    \midrule
    \multicolumn{5}{@{}l}{\textit{3D and multiview scene change detection}} \\
    3RScan~\cite{wald2019rio} & indoor rooms & re-localization & 1,482 scans & instance pose \\
    DyS2Change~\cite{Qiu20233DCL} & indoor rooms & 3D detection & 37,715 pairs & 3D box \\
    SceneDiff~\cite{Wu2025SceneDiffAB} & indoor/outdoor & 3D detection & 350 seq. pairs & instance mask \\
    \midrule
    \multicolumn{5}{@{}l}{\textit{Retail shelves (single image)}} \\
    SKU-110K~\cite{goldman2019precise} & retail & 2D detection & 11,762 images & 2D box \\
    EgoCart-OOS~\cite{allegra2021exploiting} & retail & OOS detection & 19,531 images & 2D point \\
    \midrule
    ShelfChange3D-145K & retail & 3D detection & 145K pairs & 3D box \\
    \bottomrule
  \end{tabular}
\end{table}

\subsection{ShelfChange3D-145K dataset Details}
\label{app:synth}

\subsubsection{\textbf{Assets}}
We use the Market001 scene of the GenieSim asset pack~\cite{yin2026geniesim30} in Isaac Sim~\cite{NVIDIA_Isaac_Sim}. Since the shelf assets do not expose individual boards, we split every usable board through Blender~\cite{blender_5_1_1} and record its placeable area, which defines a local frame per layer ($x$ width, $y$ depth towards the customer, $z$ up). The 67 SKUs are manually grouped by product type and shape. The product categories include personal care, beverages, snacks and packaged food, cigarettes, and household products, while the six shape classes are regular, flat, and long cuboids and regular, flat, and tall cylinders. Each SKU carries a rotational symmetry class about the vertical axis (none, 2-fold, 4-fold, continuous).

\subsubsection{\textbf{Shelf Layout Generation}}
For each shelf layer, we first sample several SKU groups. Each group is assigned an SKU and a grid size defined by its number of columns and rows (Fig.~\ref{fig:layout}a). The groups are then arranged within the valid shelf area while maintaining minimum spacing between products and groups. Products are initially placed at the centers of their grid cells, with small random position and yaw perturbations added before physics settling. Finally, the grid location of each product, represented by its group, column, and row, is recorded.

\subsubsection{\textbf{Camera Configuration}}
The simulated cameras are configured to emulate an Intel RealSense D435i. For every layer we define one or more anchor points on the board and place cameras on a sphere around each anchor, parameterized by azimuth relative to the shelf front and by elevation (Fig.~\ref{fig:cameras}a). The \emph{trajectory-level} offset is drawn once per shelf-scene and camera group. The \emph{per-frame} jitter is drawn independently every time a state is rendered, always around the fixed stopping point (Fig.~\ref{fig:cameras}b). For each pair, we store the relative pose $T_{0\to1}$ together with the camera and the relative rotation angle $\theta_{\text{rel}} = \arccos\big((\operatorname{tr}\mathbf{R}_{0\to1} - 1)/2\big)$. This rotation measure captures the full 3D orientation difference between the two cameras, including roll.

\subsubsection{\textbf{Change Generation}}
Fig.~\ref{fig:trajectory} shows one remove-add trajectory. At every removal step a few columns of the layer are chosen and the front-most product of each is taken. Additions place an SKU that does not belong to the layer's assortment either into a slot vacated at a free position on the board, with a random yaw.

\subsubsection{\textbf{Coordinate System and Format}}
All 3D annotations are expressed in the camera frame of the current observation $t_1$ (OpenCV convention), rather than in a global world frame. This allows an inspection robot to localize changed objects directly with respect to its current viewpoint, without maintaining a persistent map of the entire scene. Each box annotation further includes mask-tight and projected 2D boxes, occlusion and truncation in both frames, and SKU attributes. An example is shown in Fig.~\ref{fig:format}.

\begin{figure}[h]
  \begin{lstlisting}[basicstyle=\ttfamily\tiny]
"sample": {
  "id": 16255, "width": 640, "height": 480, "label_frame": "t1",
  "file_path": ".../frame_0004_remove/perturbed/high_left/rgb.png",
  "file_path_t1": ".../frame_0004_add/perturbed/high_left/rgb.png",
  "depth_path": "...", "depth_path_t1": "...", "depth_scale": 1000.0,
  "K_t0": [[603.0,0,319.4],[0,602.8,248.5],[0,0,1]], "K_t1": [...],
  "T_t0_to_t1": [[...],[...],[...],[0,0,0,1]], "T_t1_to_t0": [...],
  "gravity_in_cam_t0": [0.0, 0.839, 0.543], "gravity_in_cam_t1": [...],
  "view": "near_same", "composition": "add_only",
  "pair_category": "intermediate_state",
  "camera_t0": "high_left", "camera_t1": "high_left",
  "camera_geometry": {"relative_rotation_angle_deg": 2.9,
                      "baseline_m": 0.056, ...},
  "changed_removed": 0, "changed_added": 1 }
"annotation": {
  "id": 69310, "sample_id": 16255,
  "category_id": 2, "change_type": "appeared",
  "product_category": "probioticdrink002", "semantic_category": "beverage",
  "shape_category": "cylinder_regular", "symmetry": "__SYM_ROTATE_UP_4",
  "center_cam": [-0.404, 0.112, 0.899],
  "dimensions": [0.046, 0.083, 0.046],
  "pose": [[...],[...],[...]], "bbox3D_cam": [[...] x 8],
  "bbox2D_proj_t0": [9.1, 292.1, 70.3, 366.2], "bbox2D_tight_t0": null,
  "bbox2D_proj_t1": [18.2, 288.1, 76.7, 358.7], "bbox2D_tight_t1": [35, 319, 71, 353],
  "visible_px": 894,
  "occlusion_t0": 0.462, "occlusion_t1": 0.462,
  "truncation_t0": 0.0, "truncation_t1": 0.0 }
  \end{lstlisting}
  \caption{An annotation example in ShelfChange3D-145K.}
  \label{fig:format}
\end{figure}

\subsection{ShelfChange3D-5K Dataset Details}
\label{app:real}

\begin{figure}[t]
  \centering
  \includegraphics[width=\linewidth]{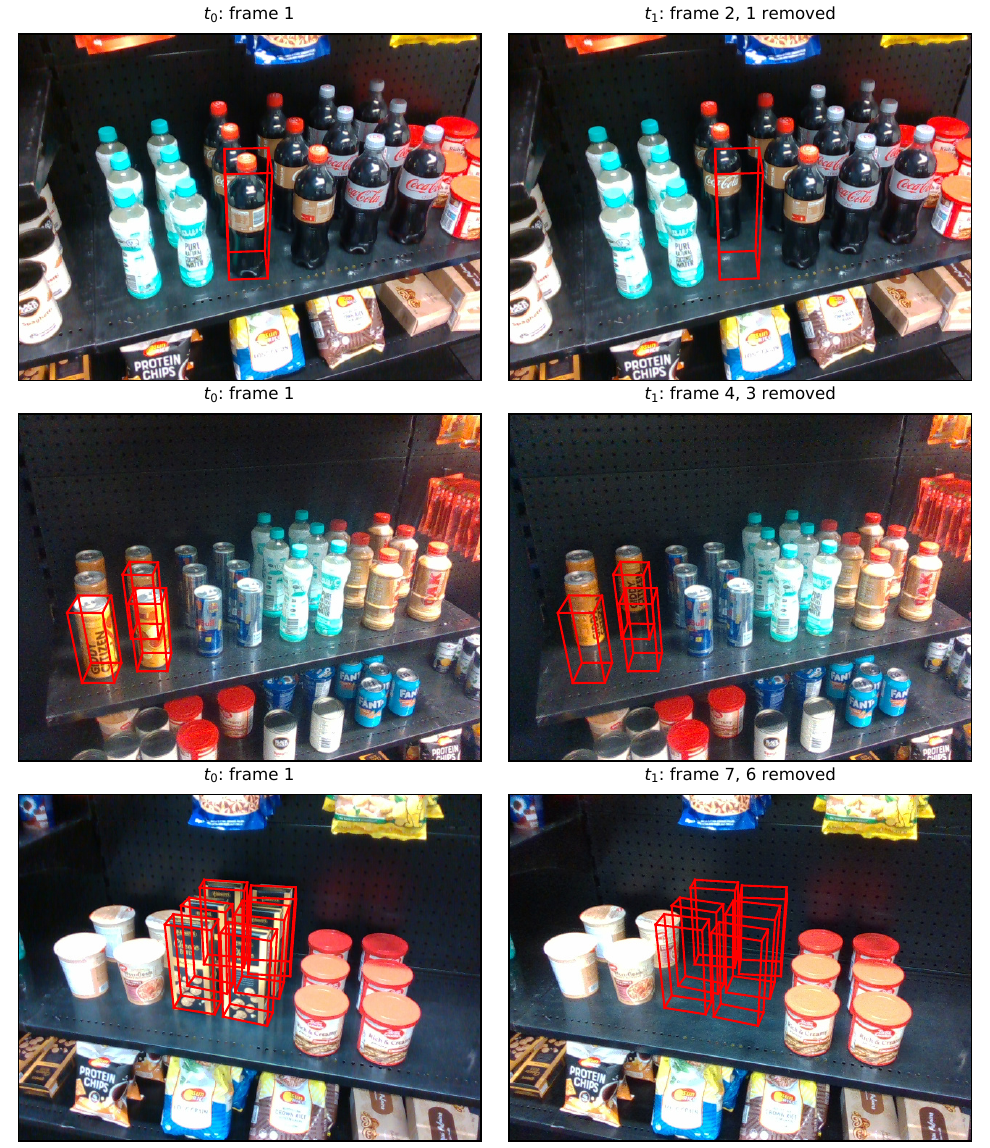}
  \caption{Examples from the ShelfChange3D-5K test split, $t_0$ on the left and $t_1$ on the right.}
  \label{fig:real_examples}
\end{figure}

Fig.~\ref{fig:real_examples} shows three examples from the real-world test split, with the reference observation $t_0$ on the left and the corresponding current observation $t_1$ on the right. The examples cover different numbers and spatial arrangements of removed products.

\subsection{Implementation Details of the Method}
\label{sec:impl}

\subsubsection{ChangeBox Detector}
\label{sec:impl_detector}

\noindent\textbf{Loss and Matching.}
In the detection loss of Eq.~\eqref{eq:loss}, the classification term is weighted by $\lambda_{cls}{=}2$ and, within the regression term of Eq.~\eqref{eq:loss_reg}, the center error by $\lambda_c{=}4$ and the size and orientation errors by $\lambda_d{=}\lambda_\theta{=}1$. The focal loss uses $\gamma{=}2$ and $\alpha{=}0.25$.

\noindent\textbf{Optimization.}
The detector is trained with the AdamW optimizer with weight decay 0.01, a batch of 24 observation pairs, bf16 mixed precision.

\noindent\textbf{Training Schedule.}
We compare three training schedules: \emph{frozen}, \emph{FT}, and
\emph{frozen$\to$FT}. For \emph{frozen}, the image backbone remains fixed throughout training,
while the remaining network is trained for 48 epochs at a learning rate of $2\times10^{-4}$
(after 500 warm-up steps), followed by 12 epochs of learning rate decay to zero.
For \emph{FT}, all parameters are optimized from the beginning using the same 48+12-epoch schedule. Our default \emph{frozen$\to$FT} schedule first follows the \emph{frozen} schedule for its 48 constant-learning-rate epochs, and then unfreezes the backbone for an additional 8 epochs, followed by 12 epochs of linear decay.

\noindent\textbf{Training on the Real Dataset.}
Fine-tuning starts from the synthetic DINOv3-S model and runs 12 epochs with a constant learning rate of $5\times10^{-5}$ followed by an 8-epoch decay. Training from scratch on the real pairs uses the frozen$\to$FT using the 48+8+12-epochs schedule.

\subsubsection{Local Box Refinement}
\label{sec:impl_refine}

\noindent\textbf{Point Collection.}
For every predicted box, the depth map of the frame in which the product is visible is sampled, and the points that fall inside the box enlarged by 3\,cm on each side are collected. At most 1{,}024 points are kept per box. A box with fewer than 30 points carries too little evidence and is returned unchanged. 

\noindent\textbf{Training Data.}
Training boxes come from two sources. \emph{Route A} (60\% of the boxes) uses real detector errors: the DINOv3-S-based detector is run on the training split and each prediction is matched to a ground-truth box of the same change type when they overlap or lie within 10\,cm of each other, keeping predictions with a score above 0.3. \emph{Route B} (40\%) uses synthetic errors: a ground-truth box is perturbed, in 45\% of the cases by an offset drawn from the residual histograms of Route A, in 10\% by stretching, in 25\% by shrinking and in 20\% by a uniform random offset. 

\noindent\textbf{Input Corruption.}
To approximate D435i depth sensing, we add Gaussian noise, edge fattening at depth discontinuities, and random depth holes. The Gaussian noise follows $\sigma(z)=1.5\,\text{mm}+2.5\,\text{mm}\cdot z^2$.

\noindent\textbf{Training and Use.}
The refinement network is trained for 16 epochs, with 24{,}000 boxes sampled per epoch, using AdamW with a batch size of 64 and an initial learning rate of $10^{-3}$ decayed with a cosine schedule. It is trained once and then used unchanged for every other backbone and for the real-world data. 

\subsubsection{DenseChangeCap Baseline}
\label{sec:impl_baseline}

\noindent\textbf{DenseChangeCap Re-implementation.}
The point-cloud baseline of Sec.~\ref{sec:exp_setup} back-projects both depth maps, builds the four clouds of DenseChangeCap~\cite{Qiu20233DCL} (before, after, before$\setminus$after, after$\setminus$before) with the differencing threshold 2\,cm. The clouds are merged with a source flag and sampled to 20,000 points in total (90\% from the two difference clouds, following the original), encodes them with PointNet++ and decodes with a 3DETR decoder. The baseline is trained from scratch on our data with the same budget as our method.

\end{document}